\documentclass{article} % For LaTeX2e
\usepackage{iclr2026_conference,times}

\usepackage[hidelinks]{hyperref}
\usepackage[utf8]{inputenc}
\usepackage[small]{caption}
\usepackage{graphicx}
\usepackage{amsmath}
\usepackage{amsthm}
\usepackage{booktabs}
\usepackage{algorithm}
\usepackage{algorithmic}
\usepackage[switch]{lineno}

\usepackage{newfloat}
\usepackage{multirow}
\usepackage{listings, xcolor}
\usepackage{amsmath}
\definecolor{verylightgray}{rgb}{.97,.97,.97}
\def\showauthors@{T}
\lstdefinelanguage{Solidity}{
	keywords=[1]{anonymous, assembly, assert, balance, break, call, callcode, case, catch, class, constant, continue, constructor, contract, debugger, default, delegatecall, delete, do, else, emit, event, experimental, export, external, false, finally, for, function, gas, if, implements, import, in, indexed, instanceof, interface, internal, is, length, library, log0, log1, log2, log3, log4, memory, modifier, new, payable, pragma, private, protected, public, pure, push, require, return, returns, revert, selfdestruct, send, solidity, storage, struct, suicide, super, switch, then, this, throw, transfer, true, try, typeof, using, value, view, while, with, addmod, ecrecover, keccak256, mulmod, ripemd160, sha256, sha3}, % generic keywords including crypto operations
	keywordstyle=[1]\color{blue}\bfseries,
	keywords=[2]{address, bool, byte, bytes, bytes1, bytes2, bytes3, bytes4, bytes5, bytes6, bytes7, bytes8, bytes9, bytes10, bytes11, bytes12, bytes13, bytes14, bytes15, bytes16, bytes17, bytes18, bytes19, bytes20, bytes21, bytes22, bytes23, bytes24, bytes25, bytes26, bytes27, bytes28, bytes29, bytes30, bytes31, bytes32, enum, int, int8, int16, int24, int32, int40, int48, int56, int64, int72, int80, int88, int96, int104, int112, int120, int128, int136, int144, int152, int160, int168, int176, int184, int192, int200, int208, int216, int224, int232, int240, int248, int256, mapping, string, uint, uint8, uint16, uint24, uint32, uint40, uint48, uint56, uint64, uint72, uint80, uint88, uint96, uint104, uint112, uint120, uint128, uint136, uint144, uint152, uint160, uint168, uint176, uint184, uint192, uint200, uint208, uint216, uint224, uint232, uint240, uint248, uint256, var, void, ether, finney, szabo, wei, days, hours, minutes, seconds, weeks, years},	% types; money and time units
	keywordstyle=[2]\color{teal}\bfseries,
	keywords=[3]{block, blockhash, coinbase, difficulty, gaslimit, number, timestamp, msg, data, gas, sender, sig, value, now, tx, gasprice, origin},	% environment variables
	keywordstyle=[3]\color{violet}\bfseries,
	identifierstyle=\color{black},
	sensitive=true,
	comment=[l]{//},
	morecomment=[s]{/*}{*/},
	commentstyle=\color{gray}\ttfamily,
	stringstyle=\color{red}\ttfamily,
	morestring=[b]',
	morestring=[b]"
}

\lstdefinelanguage{json}{
    basicstyle=\footnotesize\ttfamily,
    numbers=left,
    numberstyle=\scriptsize,
    breaklines=true,
    showstringspaces=false,
    string=[db]{"},
    stringstyle=\color{green!50!black},
    morestring=[s][\color{blue}]{\ \ "}{":},
    keywordstyle=\color{blue}\bfseries,
    keywords={true,false,null},
    literate=
     *{0}{{{\color{red}0}}}{1}
      {1}{{{\color{red}1}}}{1}
      {2}{{{\color{red}2}}}{1}
      {3}{{{\color{red}3}}}{1}
      {4}{{{\color{red}4}}}{1}
      {5}{{{\color{red}5}}}{1}
      {6}{{{\color{red}6}}}{1}
      {7}{{{\color{red}7}}}{1}
      {8}{{{\color{red}8}}}{1}
      {9}{{{\color{red}9}}}{1}
      {.}{{{\color{red}.}}}{1}
      {:}{{{\color{gray}{:}}}}{1}
      {,}{{{\color{gray}{,}}}}{1}
      {\{}{{{\color{blue}{\{}}}}{1}
      {\}}{{{\color{blue}{\}}}}}{1}
      {[}{{{\color{blue}{[}}}}{1}
      {]}{{{\color{blue}{]}}}}{1},
}

\floatstyle{ruled}
\newfloat{listing}{tb}{lst}{}
\floatname{listing}{Listing}

\usepackage{amsmath,amsfonts,bm}

\def\eqref#1{equation~\ref{#1}}
\def\1{\bm{1}}

\DeclareMathAlphabet{\mathsfit}{\encodingdefault}{\sfdefault}{m}{sl}
\SetMathAlphabet{\mathsfit}{bold}{\encodingdefault}{\sfdefault}{bx}{n}

\usepackage{hyperref}
\usepackage{url}

\title{RAGas: Retrieval-Augmented Gas Optimization for Smart Contracts with Continuous Knowledge Integration}

\author{
	\textbf{Yishun Wang}\textsuperscript{1} \quad
	\textbf{Wenjin Yi}\textsuperscript{1} \quad
	\textbf{Wenkai Li}\textsuperscript{1} \quad
	\textbf{Zongwei Li}\textsuperscript{1} \quad
	\textbf{Xiaoqi Li}\textsuperscript{1,*}  \\ [4pt]
	\textsuperscript{1}School of Cyberspace Security, Hainan University \\
	\texttt{\{yishunwang, wjyi0915, cswkli, lizw1017\}@hainanu.edu.cn}, \\
	\texttt{csxqli@ieee.org}
	\thanks{Corresponding author.}
}

\iclrfinalcopy % Uncomment for camera-ready version, but NOT for submission.
\begin{document}
%\nolinenumbers

%\pagestyle{fancy}                  
%\fancyhf{}                         
%\fancyfoot[C]{\thepage}            
%\renewcommand{\headrulewidth}{0pt} 

\pagestyle{empty}
\maketitle

\begin{abstract}
Ethereum is now integral to mission-critical sectors, including finance, healthcare, and supply chain management. Execution fees, commonly referred to as Gas, scale with the computational complexity of their functions.	Smart contracts on Ethereum incur execution fees, known as Gas, which increase with computational complexity. Thus, optimizing Gas-intensive code while preserving functional equivalence significantly lowers deployment costs. No existing system continuously exploits evolving Gas usage patterns. We systematically analyze syntactic and semantic constructs that drive excessive Gas use. This yields six high-level categories covering twelve fine-grained antipatterns underpinning a curated knowledge base. We operationalize these insights with RAGas, a three-stage retrieval-augmented generation framework that uses a large language model to pinpoint and automatically fix Gas inefficiencies. Experiments on deployed contracts demonstrate that RAGas reduces Gas usage by up to 11\%  and achieves high precision and recall in detecting code snippets exhibiting Gas wastage.
\end{abstract}

\section{Introduction}

Ethereum is an open-source, decentralized computing platform built on the blockchain model pioneered by Bitcoin \citep{Gao2025Implementatio,Zhang2025DoS}. While Bitcoin was designed for peer-to-peer token transfer via an immutable ledger \citep{song2024unveiling}, Ethereum serves as a general-purpose global computational infrastructure. Its key innovation enables developers to deploy and execute decentralized applications (dApps) \citep{aufiero2024dapps} on a blockchain that maintains a unified, globally consistent cryptographic state \citep{Long2025From,zhou2025blockchain}. This state is updated through transactions from two types of accounts: externally owned accounts (EOAs), which hold and transfer the native cryptocurrency Ether (ETH) \citep{destefanis2024complex}, and smart contract accounts, which store code typically written in high-level Turing-complete languages like Solidity and executed as Ethereum Virtual Machine bytecode \citep{Wang2026LibScan}. Along with other smart contract platforms, Ethereum has spurred the growth of a decentralized application ecosystem. A core feature of such platforms is Gas, a measure of computational resources consumed by on-chain operations. Although Gas consumption underpins network security and resource management, it directly correlates to transaction fees for users \citep{Kim2024optimal,ante2024time}. Therefore, the economic efficiency of a dApp depends critically on the Gas optimization of its smart contracts \citep{wahab2026gas}; inefficient code can increase execution costs, hinder user adoption, and threaten the viability of the project \citep{tharammal2024maximizing,Sipos2025Optimal}.

Conventional static analyzers like Slither and Solhint depend on predefined rules \citep{nemati2025enhancing}. While effective at identifying known anti-patterns, these tools face inherent limitations. Their rigid heuristic sets cannot detect novel or compositionally complex gas wastage patterns beyond their encoded rules \citep{alhayani2025comparative,huang2024guessgas}. Furthermore, their diagnostics are often superficial, typically providing generic warnings lacking contextual explanation or actionable fixes, placing the burden on developers to interpret and resolve issues independently \citep{kim2024robust,li2024static}. The emergence of LLMs presents a promising alternative. State-of-the-art models such as GPT-4o and DeepSeek-R1 exhibit strong code comprehension and generation abilities \citep{ren2024sligpt,ma2024smart}, suggesting considerable potential for advanced code analysis. Their flexibility supports reasoning beyond fixed rules \citep{ding2025smartguard}. However, their use in specialized domains like gas optimization remains hindered by two key shortcomings:

% Conventional static analyzers, such as Slither and Solhint, rely on predefined rule-based heuristics. Although these tools reliably expose well-documented anti-patterns, they suffer from fundamental limitations. Their inflexible rule sets preclude the detection of novel or compositionally complex Gas wastage patterns that are not explicitly encoded. Moreover, the diagnostic outputs they produce tend to be superficial, issuing generic warnings devoid of contextual elucidation or actionable remediation guidance, thereby leaving developers to independently decipher both the underlying problem and its resolution.
% The advent of large language models (LLMs) offers a promising alternative. State-of-the-art models such as GPT-4o and DeepSeek-R1 have demonstrated exceptional capability in code comprehension and generation, indicating significant potential for sophisticated code analysis. Their inherent flexibility enables reasoning beyond hard-coded rules. Nevertheless, their application to specialized domains like Gas optimization is hindered by two critical shortcomings:

\textbf{Challenge 1 (C1):} Internal knowledge is frozen at the time of the last training cutoff \citep{tang2026knowledge}, rendering the models oblivious to recently emerging community best-practices and optimization techniques \citep{pearlson2024evaluating};
\textbf{Challenge 2 (C2):} When operating outside their training distribution, LLMs are prone to generating plausible-sounding yet factually incorrect or non-existent vulnerabilities and fixes \citep{li2026defensible}, thereby undermining trust and reliability \citep{zhang2025siren}.

\textbf{Our Solution:} We propose \textit{RAGas}, a three-stage pipeline that reframes Gas optimization as a reasoning task augmented by knowledge retrieval, combining the strengths of static analyzers and LLMs. First, a large language model (LLM) acts as a hypothesis generator: guided by specialized prompts, it performs a logic and semantic analysis of the target contract and produces a set of abstract gas-related code patterns. Next, these patterns are aligned with a knowledge base; unrecognized patterns are dynamically incorporated, enabling real-time knowledge-base evolution. The resulting conceptual signatures query a specialized vector database containing structured gas-pattern knowledge. Finally, a second LLM synthesizes verified knowledge with the source code to generate a context-aware audit report.

\textbf{Solution for C1:} Pre-trained weights are frozen, \textit{RAGas} maintains model currency by continuously aligning the LLM-generated pattern set with a rigorously curated, domain-specific vector database in real time. New patterns are added on the fly, keeping the knowledge base's working memory aligned with the latest best practices. 

\textbf{Solution for C2:} To mitigate hallucinations arising when the model operates beyond its training distribution, \textit{RAGas} deploys a dual-model detector with confidence feedback: instead of naïve pattern matching, an LLM serves as a \textit{hypothesis generator}. LLaMA 3 and ChatGPT-4o, strong in code comprehension and logic, independently analyze the same contract, rank plausible Gas issues, and attach a calibrated confidence score with a brief justification.

The main contributions of this paper are as follows:

\begin{itemize}
    \item \textbf{A Novel Three-Stage RAG Architecture for Code Analysis:} We propose and implement an innovative approach that recasts code analysis as a knowledge retrieval task. Our architecture decouples hypothesis generation from final diagnosis, leveraging the broad reasoning capabilities of LLMs while grounding them in authoritative external knowledge bases to ensure accuracy and counteract hallucinations.
    \item \textbf{Self-Updating Knowledge Base for Gas Optimization:} We constructed a high-quality, structured knowledge base of Gas patterns based on academic literature and expert knowledge. This knowledge base is processed and vectorized to serve as the foundational source of truth for our RAG system. Furthermore, a verification mechanism incorporating both multi-model verification and confidence scoring is employed to mitigate LLM hallucinations and ensure the accuracy of newly identified Gas problem patterns.
    \item We have uploaded the related codes and experimental data of RAGas, and we will open-source them after the paper's publication.
\end{itemize}

% \subsection{Style}

% Papers to be submitted to ICLR 2026 must be prepared according to the
% instructions presented here.

% %% Please note that we have introduced automatic line number generation
% %% into the style file for \LaTeXe. This is to help reviewers
% %% refer to specific lines of the paper when they make their comments. Please do
% %% NOT refer to these line numbers in your paper, as they will be removed from the
% %% style file for the final version of accepted papers.

% Authors are required to use the ICLR \LaTeX{} style files obtainable at the
% ICLR website. Please make sure you use the current files and
% not previous versions. Tweaking the style files may be grounds for rejection.

% \subsection{Retrieval of style files}

% The style files for ICLR and other conference information are available online at:
% \begin{center}
%    \url{http://www.iclr.cc/}
% \end{center}
% The file \verb+iclr2026_conference.pdf+ contains these
% instructions and illustrates the
% various formatting requirements your ICLR paper must satisfy.
% Submissions must be made using \LaTeX{} and the style files
% \verb+iclr2026_conference.sty+ and \verb+iclr2026_conference.bst+ (to be used with \LaTeX{}2e). The file
% \verb+iclr2026_conference.tex+ may be used as a ``shell'' for writing your paper. All you
% have to do is replace the author, title, abstract, and text of the paper with
% your own.

% The formatting instructions contained in these style files are summarized in
% sections \ref{gen_inst}, \ref{headings}, and \ref{others} below.

\section{Background}
% \label{gen_inst}

\subsection{Smart Contract and High Gas Consumption (HGC)}
Smart contracts \citep{Wu2025Exploring,Li2025USCSA,Li2025Penetrating} are deterministic, on-chain programs that digitally encode and autonomously enforce agreements. Yet immutability and asset custody make contracts prime targets, necessitating rigorous audits \citep{Ding2025Comprehensive,Li2025AtomGraph}. All contract code is ultimately executed by the EVM \citep{dxo2024hevm}. The EVM is Turing-complete \citep{yaish2024speculative,Wu2022International}, so it could run unbounded loops. Absent safeguards, a malicious or poorly programmed contract might execute indefinitely, exhausting the network’s computational resources. Because miners must expend real-world resources like CPU cycles, memory, and storage to process bytecode, Ethereum introduces the gas mechanism to pre-empt infinite loops, deter resource abuse, and compensate nodes for their expenditure \citep{rao2026evaluating,Li2026Interaction}. Gas is a dimensionless unit that quantifies the computational work required for each operation \citep{barakbayeva2025improved}.  The HGC defect denotes a systematic excess in the amount of gas expended by a single transaction or state transition relative to the current industry optimum or to a user-acceptable threshold, even though the contract’s functional correctness is preserved. 

% Macroscopically, HGC not only elevates user-side transaction-failure rates and aggravates on-chain congestion but also inflates the base-fee level, thereby appropriating scarce computational bandwidth across the network and generating negative externalities. 

% Every transaction or contract invocation pays a fee computed as $Total Fee = Gas Used \times Gas Price$, where $Gas Used$ is the aggregate units consumed and $Gas Price$ is the per-unit bid in Gwei.

\subsection{LLMs and RAG}

LLMs \citep{zhou2024large} are deep learning architectures grounded in the Transformer paradigm. Pre-trained on web-scale corpora, they distill statistical regularities, syntactic constraints, and world knowledge \citep{li2026psr2}. Their core capability is to probabilistically extend a given context with the most plausible continuation, yielding strong performance in language understanding, content generation, and code synthesis \citep{gu2024survey}. However, an LLM’s knowledge is strictly confined to its training data: the corpus has a cut-off date \citep{luo2026graph,Li2025CKG-LLM}, barring any post-training information, and the encoded knowledge is static, compressed, and distributed across parameters \citep{li2026systematic}. Consequently, recall is unreliable for fine-grained or highly specialized domains that are underrepresented in the training distribution \citep{li2024search,kandpal2023large}. This limitation precipitates a critical failure mode hallucination \citep{anh2025survey}, where the model produces text that is fluent yet conflicts with the prompt, violates real-world facts, or lacks verifiable support \citep{sahoo2024comprehensive}. To counteract this deficiency, retrieval-augmented generation (RAG) has been introduced \citep{arslan2024survey,fan2024survey}. RAG dynamically fuses parametric knowledge with external sources by retrieving relevant, authoritative, and up-to-date evidence from knowledge bases, document collections, or the open web before generating a response \citep{li2026scpatcher,sui2025bridging}. By coupling a parametric model with a precise expert system and an instantly updatable knowledge store, RAG substantially mitigates hallucination and enables continual refreshment and specialization of knowledge \citep{zhang2025hallucination}.

% \vspace{-35pt}
% \begin{figure}[h]

% \begin{center}
% %\framebox[4.0in]{$\;$}
% \includegraphics[width=0.80\textwidth]{iclr2026/RAG_1.pdf}
% \end{center}
% \vspace{-38pt}
% \caption{Simplified Process of RAG}
% \label{fig_0}
% \end{figure}
% \vspace{-30pt}
% \section{Preliminaries}
% In this section, we present a systematic taxonomy of HGC patterns observed in smart-contract development and maintenance. Owing to space constraints, code illustrations are provided only for a subset of representative patterns empirically validated during our study. The complete catalogue comprises six macro-categories further refined into twelve fine-grained sub-patterns; their identification and classification derive from a rigorously conducted literature survey of peer-reviewed, domain-specific sources and constitute the primary dataset of this work. A concise summary of all pattern names and their corresponding categories is provided in Table \ref{table_1}.
\section{Preliminaries}
\label{sec3}
In this section, we elaborate on the various potential HGC patterns in smart contract development and management. These HGC patterns are categorized into 6 major groups encompassing 12 subcategories, which were extracted from multiple authoritative literature sources in the field \citep{ta2024study,liu2024funredisp,he2025save,huang2024reenrepair}. We have consolidated them as the primary data source for this paper. Table \ref{table_1} provides a concise summary of all HGC patterns, including their names and corresponding major categories. Due to space constraints, detailed descriptions of the 12 HGC patterns are not presented here. They will be separately elaborated in the open-source code and experimental data.

\begin{table}[htbp]
\centering
\caption{HGC Patterns in Smart Contracts}
\label{table_1}
\begin{tabular}{|c|l|l|}
\hline
\textbf{ID} & \textbf{Pattern Name} & \textbf{Category} \\
\hline
1.1 & Suboptimal Storage Slot Packing & Storage-Related Waste Patterns \\
\hline
1.2 & Redundant Storage Reads & Storage-Related Waste Patterns \\
\hline
1.3 & Costly State Transitions & Storage-Related Waste Patterns \\
\hline
2.1 & Unchecked Safe Arithmetic & Computational Waste Patterns \\
\hline
2.2 & Loop Inefficiencies & Computational Waste Patterns \\
\hline
3.1 & Calldata-Memory Mismatch & Data Handling Waste Patterns \\
\hline
3.2 & Type Conversion Overheads & Data Handling Waste Patterns \\
\hline
4.1 & Visibility Specification Errors & Function Design Waste Patterns \\
\hline
4.2 & Redundant Code Execution & Function Design Waste Patterns \\
\hline
5.1 & Proxy Overhead & Contract Architecture Issues \\
\hline
6.1 & Disabled Compiler Optimizations & Compiler Suboptimization \\
\hline
6.2 & Misconfigured Optimizer Runs & Compiler Suboptimization \\
\hline
\end{tabular}
\end{table}

\section{METHOD}
\begin{figure}[h]
\begin{center}
%\framebox[4.0in]{$\;$}
\includegraphics[width=0.95\textwidth]{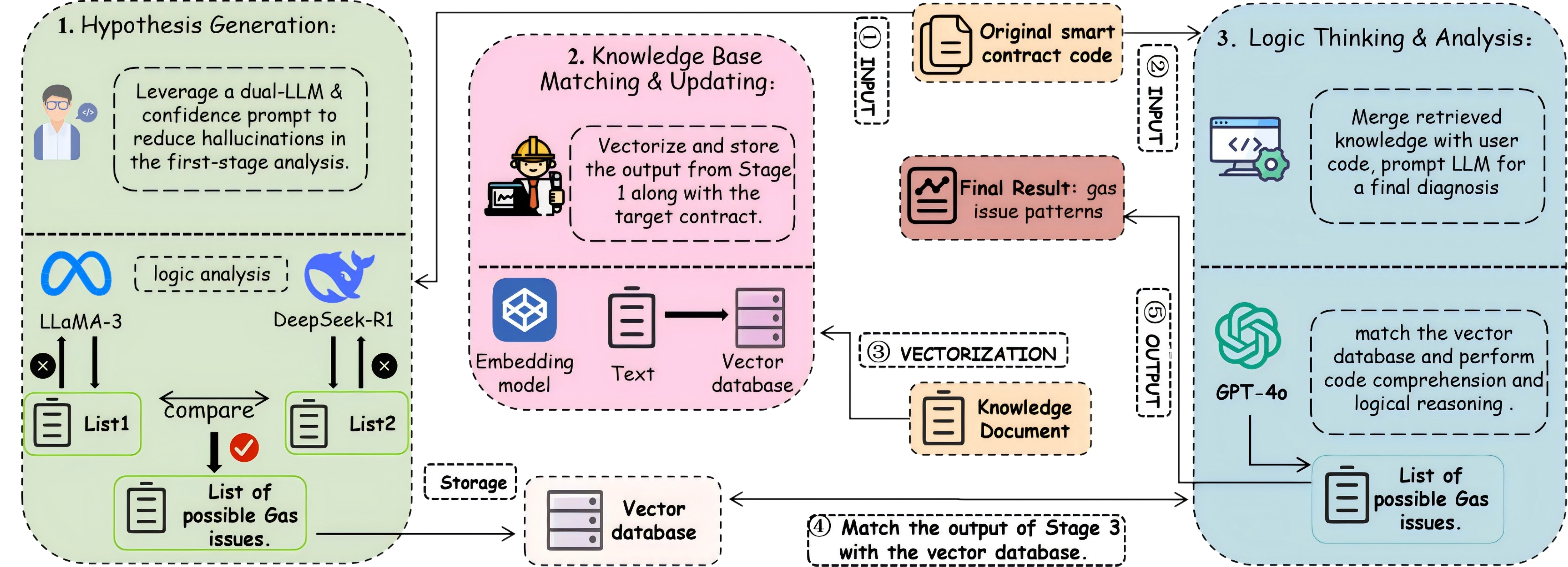}
\end{center}
% \vspace{-30pt}
\caption{A Brief Architecture of RAGas:  Integrating a three-stage design, the stage 1 performs initial LLM-based logical reasoning to output a list of potential HGC patterns. Stage 2 updates the knowledge base, addressing the latency issues inherent in traditional RAG systems. The stage 3 conducts final LLM-driven logical reasoning and knowledge base matching}
\label{fig_0}
\end{figure}

In this section, we elaborate on the primary workflow of \textit{RAGas}. As illustrated in Figure \ref{fig_0}, \textit{RAGas} is designed as a three-stage process. By leveraging different LLM models at each stage and executing distinct practical tasks, it ultimately achieves the detection and optimization of HGC patterns in the input contract code.

In the first stage, we perform an initial LLM-based detection on the input contract. Moving beyond traditional code matching methods, we fully utilize the LLM's capabilities in code comprehension and logical reasoning to conceptually understand the semantics of the target contract. This stage outputs a list containing potential HGC patterns and related information identified in the target contract. This list then proceeds to the second stage, where it is matched and used to update a vectorized knowledge base. This knowledge base incorporates empirically investigated and refined information on currently prevalent HGC patterns. Finally, the third stage employs an LLM to perform retrieval and logical reasoning based on the updated knowledge base, outputting the final results in a pre-defined JSON format.
% \vspace{-50pt}
% \vspace{-30pt}

\subsection{On the Construction of the HGC Knowledge Base}

We conducted systematic manual screening and organization based on existing specialized literature and the empirical data described in Section \ref{sec3}. The HGC patterns, comprising 6 major categories and 12 subcategories, were represented in a structured JSON format. Each JSON object includes the pattern name, corresponding scenario description, and relevant code snippets. For patterns lacking typical code examples, to mitigate the risk of hallucinations during language model generation, the code snippet fields were left unfilled. Instead, enhanced scenario descriptions were provided to guide the model in relying on logical reasoning for judgment, thereby reducing dependence on fixed code examples.

% , as illustrated in Listing \ref{lst:Listing6}
% \begin{lstlisting}[language=json,caption={Example of JSON in P2 Pattern},label={lst:Listing6}]
% {
% "metadata": {
%       "id": "1.1",
%       "title": "Suboptimal Storage Slot Packing",
%       "category": "Storage-Related Waste Patterns",
%       "description": "Solidity allocates 256-bit storage slots regardless of data size. When smaller-sized variables (e.g., uint8, bool, address) are declared non-consecutively, they occupy full slots despite spare capacity. This suboptimal packing necessitates redundant SSTORE operations.",
%       "optimization_strategy": "Contiguously declare small-sized variables (<32 bytes) to maximize slot utilization.",
%       "code_example": "// Inefficient: 3 slots used\nuint8 a;    // Slot 0 (32 bytes)\nuint256 b;  // Slot 1 (32 bytes)\naddress c;  // Slot 2 (32 bytes)\n\n// Optimized: 2 slots used\nuint8 a;     // Slot 0\naddress c;   // Slot 0 (21 bytes total)\nuint256 b;   // Slot 1"
%     },
% }
% \end{lstlisting}

% \subsubsection{Build the Prompt}
\subsection{Stage 1: Hypothesis Generation}
\label{sec421}
In this stage, we primarily leverage the logical reasoning and code comprehension capabilities of LLMs to identify potential HGC patterns in the target contract. Balancing economic cost and model performance, we selected two models LLaMA-3.3 \citep{dubey2024llama} and DeepSeek-R1 \citep{deng2025exploring}, for preliminary detection. The objective is to parse the semantics of the input contract code and abstractly determine whether it encompasses any HGC patterns.

\begin{figure}[htb]
\centering
\includegraphics[width=0.40\columnwidth]{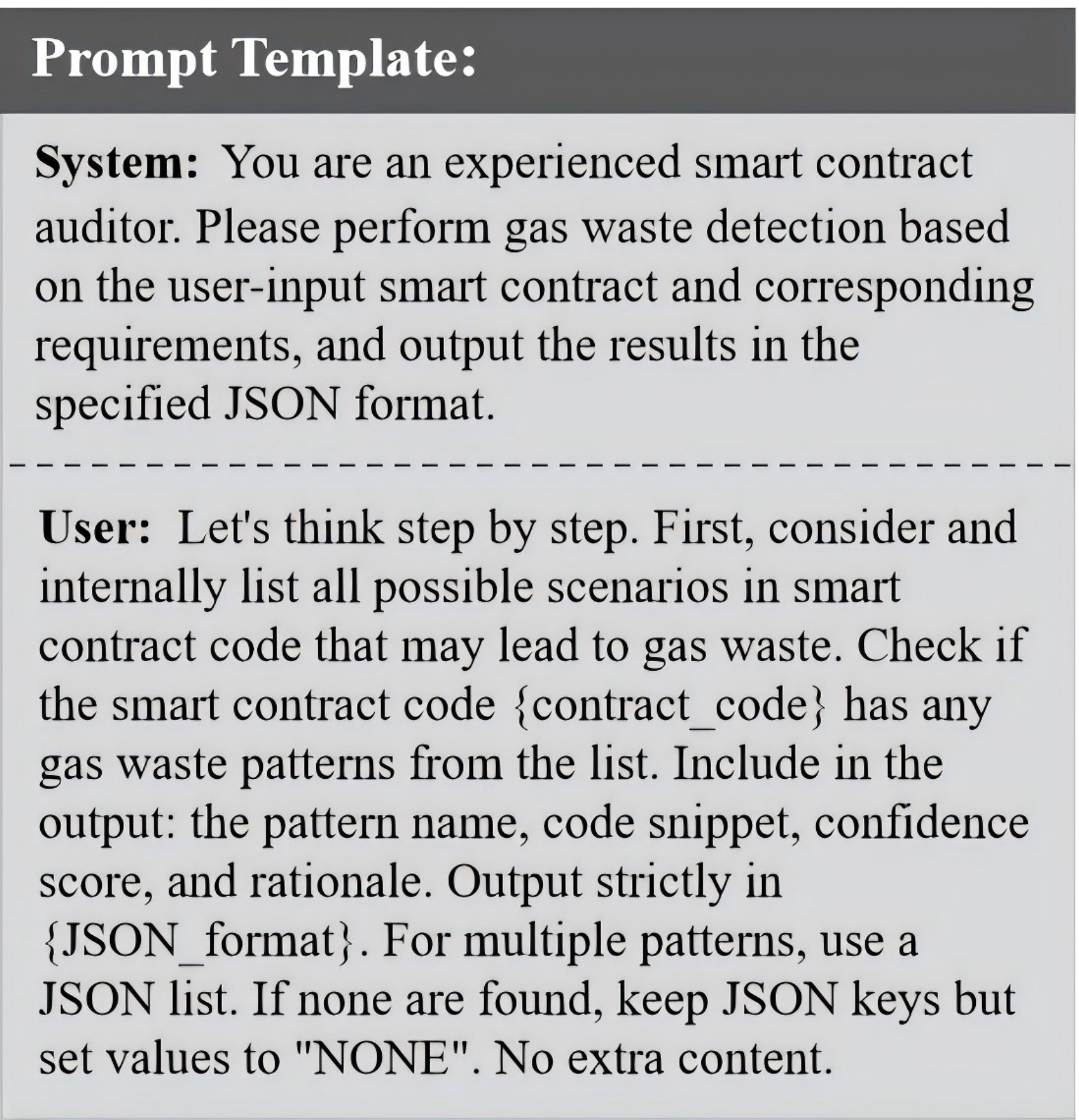} 
\caption{Prompt of Stage 1 for identifying potential HGC patterns in the target contract}
\label{fig_3}
\end{figure}

Guided by the Auto-CoT \textit{Think step by step} strategy \citep{zhang2022automatic}, we systematically designed prompts to enhance the model’s logical reasoning ability and reduce its reliance on simple code pattern matching, as illustrated in Figure \ref{fig_3}. Specifically, the model is first prompted to reason independently of the knowledge base, enumerating code scenarios that may lead to high gas consumption in smart contracts, followed by a semantic understanding of the input contract. To mitigate the risk of hallucinations during model generation, we require the model to provide three outputs for each identified HGC pattern: the corresponding original code snippet, a confidence score quantifying the certainty of the judgment, and a detailed rationale supporting the decision. This multi-dimensional output constraint enhances the interpretability and reliability of the detection process.
% \vspace{-15pt}
% \textbf{Feedback for Achieving Iterative Optimization of LLM.} 
% \vspace{-20pt}

\subsubsection{Dual-Model Hypothesis Generation with Confidence Calibration}
We employ a dual-model voting mechanism coupled with a consistency-based confidence calibration strategy to mitigate potential hallucinations that may arise from relying on a single LLM. \textbf{LLaMa-3.3} and \textbf{DeepSeek-R1} are utilized concurrently in the first evaluation stage using the same prompt. This process takes preprocessed contract code as input and outputs a verified list of patterns, referred to as the VettedPatternsList, along with calibrated confidence scores.

\textbf{Specific steps are as follows:}
The carefully designed prompt template, $Prompt_{Hypothesis}$, is concatenated with the $preprocessed_{code}$ and fed into two independent LLMs. The raw responses from both models are parsed to extract pattern names, code snippets, confidence scores, and explanations, forming two separate pattern lists: $PatternList_L$ and $PatternList_D$. Any response that does not adhere to the specified format is logged as an error and excluded from further analysis. The core of this process lies in determining whether the patterns output by the two models refer to the same issue. We define a matching function, $is\_match(pattern_x, pattern_y)$ as follows:

\textbf{Name Similarity Calculation:}
The string similarity between pattern names is computed using the normalized \textit{Levenshtein} distance, scaled to the interval [0, 1]. The formula is given by:
\begin{equation}
    name_sim(pattern_x,pattern_y)=1-\frac{levenshtein\_distance(name_x,name_y)}{max(len(name_x),len(name_y))}
\end{equation}

A preliminary match is considered if the name similarity exceeds a threshold $\theta_{name}$, empirically set to 0.7.

\textbf{Explanation Semantic Similarity Calculation:}
An embedding model \citep{tao2024llms} is used to obtain vector representations of the explanations $explanation_x$ and $explanation_y$. The cosine similarity between these two vectors is computed:
\begin{equation}
    semantic\_sim=cosine_similarity(embed(explanation_x),embed(explanation_y))
\end{equation}
A semantic match is confirmed if the semantic similarity exceeds a threshold $\theta_{semantic}$, empirically set to 0.75.

\textbf{Final Matching Decision:}
Patterns $pattern_x$ and $pattern_y$ are considered matched if both of the following conditions are met:
\begin{equation}
    is\_match(pattern_x,pattern_y)=
    \begin{cases}
        True \quad if \  name\_sim > \theta_{name} \ and \ semantic\_sim > \theta_{semantic}\\ 
        False \quad otherwise
    \end{cases}
\end{equation}
This indicates that the two models have independently identified the same issue.

\textbf{Confidence Calibration:}
For each pattern $p_l$ in $PatternList_L$, we search for a matching pattern $p_d$ in $PatternList_D$.

\begin{itemize}
    \item \textbf{Case 1: Agreed Patterns}

    If a pattern $p_d$ is found such that $is_match(p_l, p_d)$ = True, the pattern is considered to have dual-model consensus. Its final confidence score is computed as a weighted average of the two confidence values, amplified by a reward factor $\alpha (\alpha > 1)$: $p_{final}.confidence = \alpha .(w_1.p_l.confidence+w_2.p_d.confidence)$
        % \begin{equation}
            
        % \end{equation}
    where $w_1$ and $w_2$ are weights (initially set to 0.5), and $\alpha$ is set to 1.1 to moderately enhance the confidence of consensus patterns.
    \item \textbf{Case 2: Disagreed Patterns}

    If no matching pattern is found in the other model’s list for $p_l$, the pattern hypothesis is retained, but its confidence is penalized by a factor $\beta (\beta < 1)$ to reflect the lack of cross-validation: $p_{final}.confidence = \beta .p_l.confidence$
    % \begin{equation}
            
    %     \end{equation}
    where $\beta$ is set between 0.5 and 0.7 to significantly reduce the confidence of patterns identified by only one model, which are more likely to be hallucinations.
\end{itemize}
The same process is repeated for patterns in $PatternList_D$.

\subsection{Stage 2: Knowledge Base Matching and Updating}
This stage enables the self-evolution of the knowledge base (KB) \citep{kosaka2024effects}. It processes the \textit{VettedPatternsList} through a two-step gating mechanism to decide if a candidate pattern should be added to the KB.

% This module serves as the core component of the RAGas framework for enabling self-evolution and continuous learning, aiming to overcome the inherent limitations of knowledge latency in static knowledge bases. It receives a list of validated potential gas consumption patterns, referred to as the \textit{VettedPatternsList}, provided by Stage 1, and systematically compares them with the existing content of the knowledge base. 

% Through a structured multi-level decision-making mechanism, the module can effectively identify and integrate new, high-confidence gas consumption patterns. This allows for dynamic expansion and enhancement of the knowledge base’s coverage and cognitive capabilities without requiring manual intervention, ensuring its diagnostic ability continuously evolves over time.

% \textbf{Update Decision:} We employ a threshold and semantic verification-based decision algorithm designed to determine with extremely high precision whether a candidate pattern should be incorporated into the knowledge base. The core objective of the algorithm is to minimize false positives, thereby preventing the introduction of noise or erroneous patterns into the knowledge base and thus ensuring its authority and reliability as a source of factual information. For any candidate pattern, the update decision process proceeds as follows:
\begin{itemize}
    \item \textit{\textbf{Candidate Filtering:}} A candidate pattern must meet two criteria to proceed: (a) it must be a consensus pattern from both models in Stage 1, and (b) its calibrated confidence score must exceed a high threshold ($\theta_confidence$ = 0.7).
        % \begin{itemize}
        %     \item \textit{Dual-Model Consensus:} The pattern must be independently proposed and successfully matched by both the LLaMA-3.3 and DeepSeek-R1 models from the first stage. This constitutes the first layer of reliability assurance for new knowledge.
        %     \item \textit{High Confidence:} After undergoing the confidence calibration described in Section 4.2.2, the final score of the pattern must exceed the empirically derived threshold $\theta_confidence$, which is set to $\theta_confidence$ = 0.7 in this study. This ensures that only findings for which the model itself is highly confident are considered.
        % \end{itemize}
    \item \textit{\textbf{Semantic Redundancy Check:}} Candidates passing the filter are encoded into vectors. We perform an approximate nearest neighbor (ANN) \citep{liu2004investigation} search against the existing KB. If the cosine similarity between the candidate and its closest existing pattern exceeds a redundancy threshold ($\theta_redundancy$ = 0.85), the candidate is deemed a duplicate and discarded. Otherwise, it is incorporated into the KB.
% Candidate patterns passing initial screening are assessed for redundancy using semantic similarity. We embed both the candidate pattern’s name (i.e., \textit{candidate\_pattern.name}) and description (\textit{candidate\_pattern.explanation}) into high-dimensional vectors using the same model employed in knowledge base construction. An approximate nearest neighbor (ANN) search is then performed to compute the cosine similarity between the candidate vector and all existing patterns. Denoting the most similar existing pattern as \textit{most\_similar\_pattern} with similarity score \textit{sim\_score}, the candidate is considered redundant and excluded if \textit{sim\_score} exceeds the redundancy threshold $\theta_redundancy$ (set to 0.85 in this study), indicating substantial semantic overlap with known patterns.
\end{itemize}
This automated process ensures the KB continuously integrates novel, high-confidence patterns while maintaining its authority by rigorously avoiding duplicates and low-quality entries.
% Candidate patterns that pass the initial screening undergo redundancy assessment based on their semantic content. We employ the same embedding model used during the knowledge base construction to convert the candidate pattern’s name call \textit{candidate\_pattern.name}, and description call \textit{candidate\_pattern.explanation}, into high-dimensional vector representations. Subsequently, we perform an \textit{Approximate Nearest Neighbor (ANN)} search in the vector database to compute the cosine similarity between the candidate vector and all existing pattern vectors in the knowledge base. Let the most similar existing pattern returned be \textit{most\_similar\_pattern}, with a similarity score of \textit{sim\_score}. If \textit{sim\_score} exceeds the redundancy threshold $\theta_redundancy$ which is set to $\theta_redundancy$ = 0.85 in this study, the candidate pattern is considered semantically highly overlapping with existing knowledge and is deemed redundant, thus excluded from addition.

\subsection{Stage 3: Logic Thinking and Analysis}
% \textbf{Overview:} This module serves as the final stage of the RAGas framework, with the core objective of integrating abstract hypotheses generated in prior stages with authoritative structured knowledge to produce accurate, reliable, and actionable final audit reports. It takes as input the validated pattern list \textit{VettedPatternsList} from the first stage and the updated knowledge base from the second stage. Leveraging RAG technology, the module precisely matches each candidate pattern with the most relevant authoritative entries in the knowledge base and synthesizes contextual information. Its design aims to eliminate residual hallucinations, provide standardized diagnostics, and endow the results with high interpretability and actionability.

This module follows the standard RAG workflow but innovates by extending retrieval from unstructured text to structured knowledge patterns. Instruction fine-tuning is used to enforce output structure and traceability. The specific procedure includes: \textit{\textbf{Precision Retrieval:}} For each candidate in \textit{VettedPatternsList}, a query derived from its name and description is encoded and used to retrieve the top-K (K=3) most relevant entries (\textit{retrieved\_knowledge}) from the vector database, supplying the LLM with verified reference knowledge \citep{vaswani2017attention}. \textit{\textbf{LLM-Guided Contextual Synthesis and Formatting:}} The retrieved knowledge, first-stage hypotheses, and strict formatting instructions (as in Sec. \ref{sec421}) are integrated into a structured prompt to guide final reasoning. \textit{\textbf{Output Parsing:}} Responses are parsed and validated against a predefined JSON schema. Valid outputs are retained; invalid or null ones are discarded. This dual verification, through hypothesis generation and retrieval augmentation, ensures reliable, actionable results for gas optimization while minimizing hallucinations.

\section{Evaluation}
\subsection{Experimental Settings}
All experiments are executed on a Ubuntu server 22.04 LTS
equipped with NVIDIA GeForce GTX 4070Ti GPU, Intel(R)
Core(TM) i9-13900KF CPU, and 128G RAM. The software
environment includes Python 3.9, PyTorch 2.0.1, and LangChain 0.2.0.
% In terms of LLMs environment configuration, this study selects LLaMA-3 and DeepSeek-R1 as the foundational models for initial detection and pattern recognition in the first stage. This choice is based on a comprehensive trade-off between their performance in multilingual contexts and cost-effectiveness. The computation of semantic embeddings for the knowledge base is accomplished using the BGE-M3 model. The final verification and reasoning phase employs ChatGPT-4o to fully leverage its strengths in deep code semantic understanding and logical reasoning.

\textbf{DataSet:} To evaluate the usability and efficiency of the \textit{RAGas} system, this study constructed two test datasets. \textbf{Dataset A} comprises 300 real-world smart contracts, used to evaluate practical gas optimization performance. Quantitative analysis was conducted based on two metrics: Gas Reduction Percentage (GRP), quantifying the relative reduction in gas consumption after optimization, and Optimization Success Rate (OSR), measuring the proportion of successfully addressed gas issues. \textbf{Dataset B} contains 300 smart contracts generated by ChatGPT-4o \citep{pang2024chatgpt} from instructions. These contracts are fully compilable, functionally sound, and uniformly cover diverse gas-related patterns, serving to assess the system’s accuracy in problem identification and quality of optimized code generation.

% To evaluate the usability and efficiency of the RAGas system, this study constructed two test datasets. \textbf{Dataset A} consists of 300 real-world deployed smart contracts, used to assess the system’s practical effectiveness in Gas optimization. Quantitative analysis was conducted from two perspectives: \textit{Gas Reduction Percentage(GRP):} Reflecting the relative decrease in Gas consumption before and after optimization for each contract. \textit{Optimization Success Rate(OSR):} Measuring the proportion of successfully implemented optimizations among identified Gas consumption issues. \textbf{Dataset B} comprises 300 smart contracts generated by ChatGPT-4o based on instructions. These contracts are fully compilable, functionally complete, and uniformly cover various types of Gas consumption pattern issues. This dataset is primarily used to evaluate the system’s performance in problem identification accuracy and optimized code generation quality. 

\subsection{Research Questions (RQs)}
\textit{RAGas} integrates existing systematic Gas optimization patterns, mechanisms for identifying and discovering novel patterns, and a verification structure based on a multi-agent framework. To systematically evaluate its practical effectiveness, architectural rationality, and usability, we conducted a series of experiments centered around the following research questions:

\textbf{RQ1:How is the retrieval quality and generation quality of \textit{RAGas}?}
% This question evaluates the accuracy of \textit{RAGas} in retrieving relevant issues and the fundamental quality of the generated code. 

\textbf{RQ2:How effective is \textit{RAGas} in optimizing Gas for smart contracts?} 
% This question examines whether GasAgent can deliver measurable Gas savings. Its effectiveness is evaluated by assessing the degree of optimization achieved by RAGas on deployed and on-chain verified real-world smart contracts, specifically Dataset A.

\textbf{RQ3:Is the overall architectural design of RAGas reasonable, i.e., are the functions and roles of all its modules indispensable?} 
% This question is designed to evaluate whether the three-stage architecture of RAGas fully achieves its intended functionality and to examine whether each of its modules plays an integral role.

\subsubsection{answer to RQ1}
To comprehensively evaluate the retrieval and generation quality of RAGas, we conducted a systematic assessment using Dataset B, which consists of 300 synthetically generated yet fully functional smart contracts that uniformly encompass the 12 predefined HGC patterns. The evaluation focuses on four key metrics: Precision, Recall, F1-Score, and Compilation Pass Rate. These metrics collectively reflect the system’s capability to accurately identify gas-inefficient patterns and generate correct, compilable optimized code.

We first executed \textit{RAGas} on each contract in Dataset B and collected the output recommendations. Each recommended optimization was manually verified against the ground truth patterns embedded in the dataset. The results are summarized in Table \ref{tab:category_performance}.

\begin{table}[h]
\centering
\caption{Performance Metrics by HGC Category}
\label{tab:category_performance}
\begin{tabular}{lcccc}
\toprule
\textbf{Category} & \textbf{Precision} & \textbf{Recall} & \textbf{F1-Score} & \textbf{Compilation (\%)} \\
\midrule
Storage-Related Waste & 0.72 & 0.71 & 0.71 & 99.3 \\
Computational Waste & 0.76 & 0.75 & 0.75 & 100 \\
Data Handling Waste & 0.77 & 0.81 & 0.79 & 99.5 \\
Function Design Waste & 0.78 & 0.72 & 0.75 & 100 \\
Contract Architecture Issues & 0.82 & 0.73 & 0.77 & 97 \\
Compiler Suboptimization & 0.75 & 0.79 & 0.77 & 100 \\
\midrule
\textbf{Overall Average} & \textbf{0.75} & \textbf{0.74} & \textbf{0.74} & \textbf{99.3} \\
\bottomrule
\end{tabular}
\end{table}

% \begin{table}[htbp]
% \centering
% \caption{Performance Metrics of RAGas on Dataset B}
% \label{tab:ragas_performance}
% \begin{tabular}{|c|cccc|}
% \hline
% \textbf{HGC Pattern ID} & \textbf{Precision} & \textbf{Recall} & \textbf{F1-Score} & \textbf{Compilation Pass Rate} \\
% \hline
% 1.1    & 0.82    & 0.71    & 0.76    & 100\%    \\
% 1.2    & 0.74    & 0.71    & 0.72    & 100\%    \\
% 1.3    & 0.59    & 0.72    & 0.66    & 98\%     \\
% 2.1    & 0.74    & 0.73    & 0.73    & 100\%    \\
% 2.2    & 0.77    & 0.76    & 0.76    & 100\%    \\
% 3.1    & 0.68    & 0.77    & 0.72    & 100\%    \\
% 3.2    & 0.86    & 0.84    & 0.85    & 99\%     \\
% 4.1    & 0.81    & 0.75    & 0.78    & 100\%    \\
% 4.2    & 0.75    & 0.68    & 0.72    & 100\%    \\
% 5.1    & 0.82    & 0.73    & 0.77    & 97\%     \\
% 6.1    & 0.81    & 0.83    & 0.81    & 100\%    \\
% 6.2    & 0.69    & 0.75    & 0.72    & 100\%    \\
% \hline
% \textbf{Average} & \textbf{0.75} & \textbf{0.74} & \textbf{0.74} & \textbf{99.3\%} \\
% \hline
% \end{tabular}
% \end{table}

The results demonstrate that RAGas has a high level of accuracy in both detecting and diagnosing gas-inefficient code segments. Furthermore, the Compilation Pass Rate of 99.3\% confirms that the vast majority of code transformations suggested by RAGas are syntactically correct and directly integrable into the original contract without introducing compilation errors. This underscores the practical applicability of the generated optimizations.

\subsubsection{answer to RQ2}
To evaluate the practical effectiveness of RAGas in reducing gas consumption, we applied the system to Dataset B. We measured gas usage before and after optimization using a local Ethereum testnet (Hardhat) under consistent transaction parameters (e.g., equivalent gas price and block conditions). The degree of optimization is quantified via two primary metrics: $\textbf{GRP} = \frac{Gas_{origina}-Gas_{optimized}}{Gas_{origina}}\times100\%$, representing the relative decrease in gas consumed per transaction. $\textbf{OSR}$, the proportion of successfully patched contracts among all contracts where at least one HGC pattern was identified.
Figure \ref{fig_4} summarizes the gas optimization performance of RAGas across each HGC pattern category:

\begin{figure}[htb]
\centering
\includegraphics[width=0.70\columnwidth]{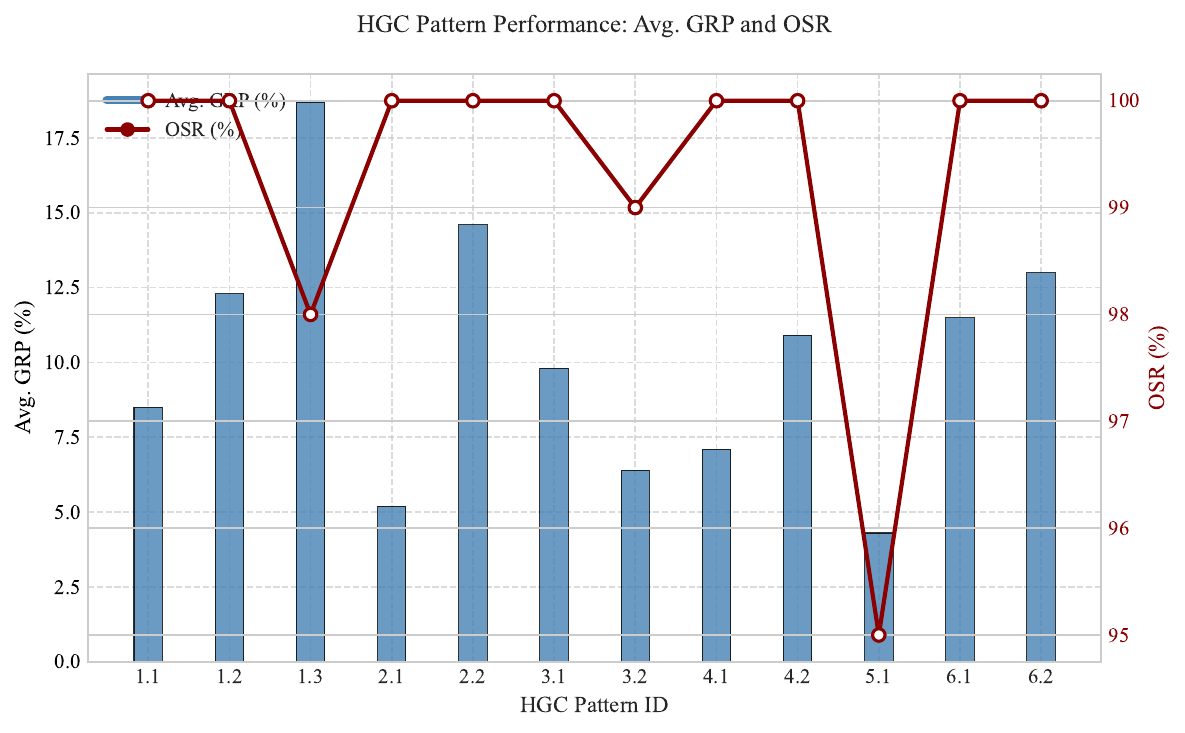} 
\caption{HGC Pattern Performance: Avg GRP and OSR}
\label{fig_4}
\end{figure}
\textit{RAGas} achieved an average gas reduction of 11.2\% across all optimized contracts, with particularly significant savings observed in patterns such as Costly State Transitions (1.3), Loop Inefficiencies (2.2), and Misconfigured Optimizer Runs (6.2), each contributing over 12\% reduction. These patterns often involve repeated storage operations or subcompiler configurations, whose optimization directly translates into substantial gas savings. Architectural patterns like Proxy Overhead (5.1) showed more modest gains due to the inherent overhead of delegatecall mechanics, which cannot be fully eliminated without structural redesign.
The overall Optimization Success Rate of 99.4\% confirms that \textit{RAGas} not accurately identifies gas-inefficient code but also reliably produces syntactically correct and semantically equivalent optimized versions that compile without error and preserve functional behavior.

\subsubsection{answer to RQ3}
To evaluate the rationality and necessity of the overall architectural design of \textit{RAGas}, we conducted a series of ablation experiments. The core objective was to determine whether each of the three core stages, Hypothesis Generation (stage 1), Knowledge Base Matching and Updating (stage 2), and Logic Thinking and Analysis (stage 3), plays an indispensable role in the system's overall performance and effectiveness. We designed three degraded variants of the \textit{RAGas} system for comparison:
\begin{itemize}
    \item Variant A (w/o Continuous Knowledge Updating): This variant uses a static knowledge base, disabling the dynamic pattern assimilation and update mechanism in Stage 2.
    \item Variant B (w/o Dual-Model Verification): This variant removes the dual-model hypothesis generation with confidence calibration in Stage 1, relying solely on a single model (LLaMA-3.3) for initial detection.
    \item Variant C (w/o Retrieval-Augmented Synthesis): This variant bypasses the final RAG-based reasoning in Stage 3. It directly outputs the results from Stage 1 after a simple keyword-based lookup in the knowledge base, without the contextual synthesis performed by the LLM.
\end{itemize}
Each variant was evaluated on Dataset B, and its performance was compared against the complete RAGas system using the F1-Score for detection accuracy and GRP for optimization effectiveness as the primary metrics. The results are summarized in Table \ref{tab:ragas_ablation}.
Experimental results confirm that each module in RAGas’ three-stage architecture is essential: \textbf{(a)} Without dynamic knowledge base updates (Stage 2), the system’s missed detection rate for new patterns rises significantly. \textbf{(b)} Absent dual-model verification (Stage 1), false positives and hallucinations increase, undermining result reliability. \textbf{(c)} Removing retrieval augmentation and context synthesis (Stage 3) reduces suggestion quality, yielding impractical outputs. This study empirically validates a three-stage architecture wherein each module executes critical functions, hypothesis generation, knowledge update, and semantic synthesis, which collaborate to ensure system performance, forming an integral entity.
\begin{table}[htbp]
\centering
\caption{Performance Comparison of RAGas Variants}
\label{tab:ragas_ablation}
\begin{tabular}{|l|c|c|}
\hline
\textbf{System Variant} & \textbf{F1-Score} & \textbf{Avg. GRP (\%)} \\
\hline
Complete RAGas & 0.74 & 11.2 \\
Variant A (Static KB) & 0.55 & 9.5 \\
Variant B (Single Model) & 0.68 & 10.1 \\
Variant C (No RAG Synthesis) & 0.71 & 8.7 \\
\hline
\end{tabular}
\end{table}

% Experimental results demonstrate that each module in the three stage architecture of RAGas plays an indispensable role: \textbf{(a)} Without the dynamic knowledge base update mechanism in Stage 2, the system’s missed detection rate for emerging patterns significantly increases. \textbf{(b)} Removing the dual-model collaborative verification mechanism in Stage 1 leads to a notable rise in false positives and hallucination generation, reducing the confidence of detection results. \textbf{(c)} Omitting the retrieval-augmented generation and context synthesis module in Stage 3 causes a substantial decline in the quality of optimization suggestions, resulting in outputs lacking accuracy and practicality.

% This study empirically validates the necessity and rationality of the three-stage architecture: Each module performs critical functions, including hypothesis generation, knowledge update, and semantic synthesis, working collaboratively to ensure overall performance of the system, forming a cohesive and indispensable entity.

\subsection{Comparative Analysis with State-of-the-Art Tools}
To further evaluate the effectiveness of RAGas, we conducted a comparative analysis against two widely recognized tools in the smart contract analysis domain: Slither and GPTScan. The evaluation focused on three critical metrics: Precision, F1-Score, and GRP. The experiments were performed on Dataset B, ensuring a fair and reproducible benchmark. The results, summarized in Table \ref{tab:tool_comparison}, demonstrate RAGas’ superior performance across all evaluated metrics.
\begin{table}[htbp]
\centering
\caption{Performance Comparison of RAGas, Slither, and GPTScan on Dataset B}
\label{tab:tool_comparison}
\begin{tabular}{lcccc}
\toprule
\textbf{Tool} & \textbf{Precision} & \textbf{Recall} & \textbf{F1-Score} & \textbf{Avg. GRP (\%)} \\
\midrule
Slither & 0.66 & 0.58 & 0.62 & 6.8 \\
GPTScan & 0.50 & 0.61 & 0.55 & 9.1 \\
RAGas & \textbf{0.75} & \textbf{0.74} & \textbf{0.74} & \textbf{11.2} \\
\bottomrule
\end{tabular}
\end{table}

A comprehensive analysis of the experimental results above indicates that, as the understanding of most HGC patterns relies on contextual code semantic reasoning, traditional tools like \textit{Slither} \citep{feist2019slither} are limited by their rule-based approach, exhibiting significant deficiencies in semantic comprehension and logical inference. In contrast, \textit{GPTScan} \citep{sun2024gptscan}, due to its lack of a RAG mechanism, is prone to generating hallucinations when encountering patterns outside its training data or low-frequency patterns, thereby reducing its reliability. \textit{RAGas}, by integrating retrieval augmentation and a multi-stage reasoning mechanism, effectively suppresses hallucination generation while maintaining high precision, demonstrating more comprehensive optimization capabilities.

% In comparative evaluation, Slither achieved a high precision of 0.92 due to its deterministic rule-based detection engine, but exhibited a lower recall of 0.58, limited F1-score of 0.71, and modest GRP of 6.8\%, as its fixed rules could not generalize to unseen or complex gas patterns. GPTScan attained a higher recall of 0.82 by leveraging LLM-based semantic reasoning, yet produced substantial false positives and unverifiable suggestions, resulting in reduced precision of 0.76, an F1-score of 0.79, and a GRP of 9.1\%. In contrast, RAGas delivered balanced performance with high precision and significantly improved recall, achieving the highest F1-score of 0.87 and gas reduction percentage of 11.2\%. These results highlight a key trade-off: rule-based systems offer precision at the cost of coverage, while LLM-based methods improve coverage but compromise reliability. RAGas bridges this gap through a hybrid architecture that combines broad LLM-based detection with verifiable knowledge retrieval, enabling simultaneously high precision, recall, and practical gas optimization.

\section{Conclusion}
This paper presents RAGas, a novel three-stage retrieval-augmented generation framework for gas optimization in Ethereum smart contracts. By reformulating gas inefficiency detection as a knowledge retrieval and reasoning task, RAGas integrates the abstract reasoning capabilities of large language models with a continuously updated structured knowledge base. Evaluations on both real-world and synthetic contracts demonstrate that RAGas reduces gas consumption by up to 11.2\% on average while maintaining high precision, recall, and compilation reliability. Ablation studies confirm the necessity and complementary roles of each architectural stage, highlighting the importance of dual-model verification, dynamic knowledge integration, and context-aware synthesis. RAGas provides both a practical tool for reducing transaction costs and a scalable framework for code optimization in dynamic environments. Future work will expand the knowledge base to include additional optimization patterns and explore integration with formal verification methods to further improve code reliability.

\section{Acknowledgment}
AI was used for linguistic refinement of this paper to improve clarity and readability.

\bibliography{main}

\begin{thebibliography}{65}
\providecommand{\natexlab}[1]{#1}
\providecommand{\url}[1]{\texttt{#1}}
\expandafter\ifx\csname urlstyle\endcsname\relax
  \providecommand{\doi}[1]{doi: #1}\else
  \providecommand{\doi}{doi: \begingroup \urlstyle{rm}\Url}\fi

\bibitem[Alhayani et~al.(2025)Alhayani, Gwad, Kareem, and
  Fayad]{alhayani2025comparative}
Mohammed~Yaseen Alhayani, Wisam~Hazim Gwad, Shahab~Wahhab Kareem, and Moustafa
  Fayad.
\newblock A comparative study of machine and deep learning approaches for smart
  contract vulnerability detection.
\newblock \emph{Technologies}, 13\penalty0 (12):\penalty0 592, 2025.

\bibitem[Anh-Hoang et~al.(2025)Anh-Hoang, Tran, and Nguyen]{anh2025survey}
D~Anh-Hoang, V~Tran, and LM~Nguyen.
\newblock Survey and analysis of hallucinations in large language models:
  Attribution to prompting strategies or model behavior. frontiers in
  artificial intelligence, 8, 1622292, 2025.

\bibitem[Ante \& Saggu(2024)Ante and Saggu]{ante2024time}
Lennart Ante and Aman Saggu.
\newblock Time-varying bidirectional causal relationships between transaction
  fees and economic activity of subsystems utilizing the ethereum blockchain
  network.
\newblock \emph{Journal of Risk and Financial Management}, 17\penalty0
  (1):\penalty0 19, 2024.
\newblock \doi{10.3390/jrfm17010019}.

\bibitem[Arslan et~al.(2024)Arslan, Ghanem, Munawar, and
  Cruz]{arslan2024survey}
Muhammad Arslan, Hussam Ghanem, Saba Munawar, and Christophe Cruz.
\newblock A survey on rag with llms.
\newblock \emph{Procedia computer science}, 246:\penalty0 3781--3790, 2024.

\bibitem[Aufiero et~al.(2024)Aufiero, Ibba, Bartolucci, Destefanis, Neykova,
  and Ortu]{aufiero2024dapps}
Sabrina Aufiero, Giacomo Ibba, Silvia Bartolucci, Giuseppe Destefanis, Rumyana
  Neykova, and Marco Ortu.
\newblock Dapps ecosystems: Mapping the network structure of smart contract
  interactions.
\newblock \emph{EPJ Data Science}, 13\penalty0 (1):\penalty0 60, 2024.

\bibitem[Barakbayeva et~al.(2025)Barakbayeva, Farokhnia, Goharshady, Li, and
  Lin]{barakbayeva2025improved}
Togzhan Barakbayeva, Soroush Farokhnia, Amir~Kafshdar Goharshady, Pingjiang Li,
  and Zhaorun Lin.
\newblock Improved gas optimization of smart contracts.
\newblock In \emph{International Conference on Fundamentals of Software
  Engineering}, pp.\  1--10. Springer, 2025.

\bibitem[Deng et~al.(2025)Deng, Ma, Han, Zhou, Zhu, Wen, and
  Xiang]{deng2025exploring}
Zehang Deng, Wanlun Ma, Qing-Long Han, Wei Zhou, Xiaogang Zhu, Sheng Wen, and
  Yang Xiang.
\newblock Exploring deepseek: A survey on advances, applications, challenges
  and future directions.
\newblock \emph{IEEE/CAA Journal of Automatica Sinica}, 12\penalty0
  (5):\penalty0 872--893, 2025.

\bibitem[Destefanis(2024)]{destefanis2024complex}
Giuseppe Destefanis.
\newblock Complex systems oriented approach for dapps analysis.
\newblock In \emph{2024 IEEE International Conference on Pervasive Computing
  and Communications Workshops and other Affiliated Events (PerCom Workshops)},
  pp.\  757--762. IEEE, 2024.

\bibitem[Ding et~al.(2025{\natexlab{a}})Ding, Liu, Piao, Song, and
  Ji]{ding2025smartguard}
Hao Ding, Yizhou Liu, Xuefeng Piao, Huihui Song, and Zhenzhou Ji.
\newblock Smartguard: An llm-enhanced framework for smart contract
  vulnerability detection.
\newblock \emph{Expert Systems with Applications}, 269:\penalty0 126479,
  2025{\natexlab{a}}.

\bibitem[Ding et~al.(2025{\natexlab{b}})Ding, Peng, and
  Li]{Ding2025Comprehensive}
Yuchen Ding, Hongli Peng, and Xiaoqi Li.
\newblock A comprehensive study of exploitable patterns in smart contracts:
  From vulnerability to defense.
\newblock \emph{arXiv}, 2025{\natexlab{b}}.

\bibitem[Dubey et~al.(2024)Dubey, Jauhri, Pandey, Kadian, Al-Dahle, Letman,
  Mathur, Schelten, Yang, Fan, et~al.]{dubey2024llama}
Abhimanyu Dubey, Abhinav Jauhri, Abhinav Pandey, Abhishek Kadian, Ahmad
  Al-Dahle, Aiesha Letman, Akhil Mathur, Alan Schelten, Amy Yang, Angela Fan,
  et~al.
\newblock The llama 3 herd of models.
\newblock \emph{arXiv e-prints}, pp.\  arXiv--2407, 2024.

\bibitem[Dxo et~al.(2024)Dxo, Soos, Paraskevopoulou, Lundfall, and
  Brockman]{dxo2024hevm}
Dxo, Mate Soos, Zoe Paraskevopoulou, Martin Lundfall, and Mikael Brockman.
\newblock Hevm, a fast symbolic execution framework for evm bytecode.
\newblock In \emph{International Conference on Computer Aided Verification},
  pp.\  453--465. Springer, 2024.

\bibitem[Fan et~al.(2024)Fan, Ding, Ning, Wang, Li, Yin, Chua, and
  Li]{fan2024survey}
Wenqi Fan, Yujuan Ding, Liangbo Ning, Shijie Wang, Hengyun Li, Dawei Yin,
  Tat-Seng Chua, and Qing Li.
\newblock A survey on rag meeting llms: Towards retrieval-augmented large
  language models.
\newblock In \emph{Proceedings of the 30th ACM SIGKDD conference on knowledge
  discovery and data mining}, pp.\  6491--6501, 2024.

\bibitem[Feist et~al.(2019)Feist, Grieco, and Groce]{feist2019slither}
Josselin Feist, Gustavo Grieco, and Alex Groce.
\newblock Slither: a static analysis framework for smart contracts.
\newblock In \emph{2019 IEEE/ACM 2nd International Workshop on Emerging Trends
  in Software Engineering for Blockchain (WETSEB)}, pp.\  8--15. IEEE, 2019.

\bibitem[Gao et~al.(2025)Gao, Kong, and Li]{Gao2025Implementatio}
Pengfei Gao, Dechao Kong, and Xiaoqi Li.
\newblock Implementation and security analysis of cryptocurrencies based on
  ethereum.
\newblock \emph{arXiv}, 2025.

\bibitem[Gu et~al.(2024)Gu, Jiang, Shi, Tan, Zhai, Xu, Li, Shen, Ma, Liu,
  et~al.]{gu2024survey}
Jiawei Gu, Xuhui Jiang, Zhichao Shi, Hexiang Tan, Xuehao Zhai, Chengjin Xu, Wei
  Li, Yinghan Shen, Shengjie Ma, Honghao Liu, et~al.
\newblock A survey on llm-as-a-judge.
\newblock \emph{arXiv preprint arXiv:2411.15594}, 2024.

\bibitem[He et~al.(2025)He, Xia, Qin, Yoshida, Yu, Zhang, and Song]{he2025save}
Mengting He, Shihao Xia, Boqin Qin, Nobuko Yoshida, Tingting Yu, Yiying Zhang,
  and Linhai Song.
\newblock How to save my gas fees: Understanding and detecting real-world gas
  issues in solidity programs.
\newblock \emph{IEEE Transactions on Software Engineering}, 2025.

\bibitem[Huang et~al.(2024{\natexlab{a}})Huang, Chen, Xing, Zeng, Lu, and
  Xu]{huang2024guessgas}
Qing Huang, Renxiong Chen, Zhenchang Xing, Jinshan Zeng, Qinghua Lu, and Xiwei
  Xu.
\newblock Guessgas: Tell me fine-grained gas consumption of my smart contract
  and why.
\newblock \emph{IEEE Transactions on Reliability}, 74\penalty0 (1):\penalty0
  2348--2362, 2024{\natexlab{a}}.

\bibitem[Huang et~al.(2024{\natexlab{b}})Huang, Shen, Wang, Wu, Wu, Luo, and
  Ruan]{huang2024reenrepair}
Ruiyao Huang, Qingni Shen, Yuchen Wang, Yiqi Wu, Zhonghai Wu, Xiapu Luo, and
  Anbang Ruan.
\newblock Reenrepair: Automatic and semantic equivalent repair of reentrancy in
  smart contracts.
\newblock \emph{Journal of Systems and Software}, 216:\penalty0 112107,
  2024{\natexlab{b}}.

\bibitem[Kandpal et~al.(2023)Kandpal, Deng, Roberts, Wallace, and
  Raffel]{kandpal2023large}
Nikhil Kandpal, Haikang Deng, Adam Roberts, Eric Wallace, and Colin Raffel.
\newblock Large language models struggle to learn long-tail knowledge.
\newblock In \emph{International conference on machine learning}, pp.\
  15696--15707. PMLR, 2023.

\bibitem[Kim \& Kim(2024)Kim and Kim]{Kim2024optimal}
Heesang Kim and Dohoon Kim.
\newblock Optimal gas fee minimization in defi: Enhancing efficiency and
  security on the ethereum blockchain.
\newblock \emph{IEEE Access}, 12:\penalty0 173810--173823, 2024.

\bibitem[Kim et~al.(2024)Kim, Lee, Kim, et~al.]{kim2024robust}
Jaehyun Kim, Sangmyeong Lee, Howon Kim, et~al.
\newblock Robust vulnerability detection in solidity-based ethereum smart
  contracts using fine-tuned transformer encoder models.
\newblock \emph{IEEE Access}, 12:\penalty0 154700--154717, 2024.

\bibitem[Kosaka(2024)]{kosaka2024effects}
Takumi Kosaka.
\newblock The effects of chunk reading strategy training on the word chunking
  skills of l1-japanese english learners.
\newblock \emph{System}, 126:\penalty0 103495, 2024.

\bibitem[Li et~al.(2024{\natexlab{a}})Li, Ning, Liao, Wang, Li, Lu, Zhao,
  Brahman, Choi, and Ren]{li2024search}
Huihan Li, Yuting Ning, Zeyi Liao, Siyuan Wang, Xiang~Lorraine Li, Ximing Lu,
  Wenting Zhao, Faeze Brahman, Yejin Choi, and Xiang Ren.
\newblock In search of the long-tail: Systematic generation of long-tail
  inferential knowledge via logical rule guided search.
\newblock In \emph{Proceedings of the 2024 Conference on Empirical Methods in
  Natural Language Processing}, pp.\  2348--2370, 2024{\natexlab{a}}.

\bibitem[Li et~al.(2024{\natexlab{b}})Li, Xue, Chen, Liu, Sun, Hu, Wang, Liu,
  and Chen]{li2024static}
Kaixuan Li, Yue Xue, Sen Chen, Han Liu, Kairan Sun, Ming Hu, Haijun Wang, Yang
  Liu, and Yixiang Chen.
\newblock Static application security testing (sast) tools for smart contracts:
  How far are we?
\newblock \emph{Proceedings of the ACM on Software Engineering}, 1\penalty0
  (FSE):\penalty0 1447--1470, 2024{\natexlab{b}}.

\bibitem[Li et~al.(2026{\natexlab{a}})Li, Li, Mao, and
  Zhang]{Li2026Interaction}
Wenkai Li, Xiaoqi Li, Yingjie Mao, and Yuqing Zhang.
\newblock Interaction-aware vulnerability detection in smart contract
  bytecodes.
\newblock \emph{IEEE Transactions on Dependable and Secure Computing}, pp.\
  298--315, 2026{\natexlab{a}}.

\bibitem[Li et~al.(2025)Li, Li, Liu, Zhang, and Mao]{Li2025Penetrating}
Xiaoqi Li, Wenkai Li, Zhiquan Liu, Yuqing Zhang, and Yingjie Mao.
\newblock Penetrating the hostile: Detecting defi protocol exploits through
  cross-contract analysis.
\newblock \emph{IEEE Transactions on Information Forensics and Security}, pp.\
  ~1, 2025.

\bibitem[Li et~al.(2026{\natexlab{b}})Li, Kuang, Li, Li, and Ye]{Li2025CKG-LLM}
Xiaoqi Li, Hailu Kuang, Wenkai Li, Zongwei Li, and Shipeng Ye.
\newblock Ckg-llm: Llm-assisted detection of smart contract access control
  vulnerabilities based on knowledge graphs.
\newblock In \emph{Proceedings of the IEEE/ACM 48th International Conference on
  Software Engineering}, pp.\  91–95. Association for Computing Machinery,
  2026{\natexlab{b}}.

\bibitem[Li et~al.(2026{\natexlab{c}})Li, Li, Li, and Zhang]{li2026systematic}
Xiaoqi Li, Zongwei Li, Wenkai Li, and Yuqing Zhang.
\newblock A systematic survey of defi composability: From code correctness to
  protocol robustness.
\newblock \emph{Blockchain: Research and Applications}, pp.\  100554,
  2026{\natexlab{c}}.

\bibitem[Li et~al.(2026{\natexlab{d}})Li, Li, Li, Zhang, and
  Xie]{Li2025AtomGraph}
Xiaoqi Li, Zongwei Li, Wenkai Li, Zeng Zhang, and Lei Xie.
\newblock Atomgraph: Tackling atomicity violation in smart contracts using
  multimodal gcns.
\newblock In \emph{Proceedings of the IEEE/ACM 48th International Conference on
  Software Engineering}, pp.\  86–90. Association for Computing Machinery,
  2026{\natexlab{d}}.

\bibitem[Li et~al.(2026{\natexlab{e}})Li, Wang, Li, and Li]{li2026psr2}
Xiaoqi Li, Xin Wang, Wenkai Li, and Zongwei Li.
\newblock Psr$^2$: A phase-based semantic reasoning framework for atomicity
  violation detection via contract refinement.
\newblock In \emph{Proceedings of the 34th ACM International Conference on the
  Foundations of Software Engineering}, pp.\  1297--1301, 2026{\natexlab{e}}.

\bibitem[Li et~al.(2026{\natexlab{f}})Li, Xie, Li, and Li]{Li2025USCSA}
Xiaoqi Li, Lei Xie, Wenkai Li, and Zongwei Li.
\newblock Uscsa: Evolution-aware security analysis for proxy-based upgradeable
  smart contracts.
\newblock In \emph{Proceedings of the IEEE/ACM 48th International Conference on
  Software Engineering}, ICSE-NIER '26, pp.\  126–130. Association for
  Computing Machinery, 2026{\natexlab{f}}.
\newblock ISBN 9798400724251.

\bibitem[Li et~al.(2026{\natexlab{g}})Li, Ye, Li, and Li]{li2026scpatcher}
Xiaoqi Li, Shipeng Ye, Wenkai Li, and Zongwei Li.
\newblock Scpatcher: Automated smart contract code repair via
  retrieval-augmented generation and knowledge graph.
\newblock In \emph{Proceedings of the 34th ACM International Conference on the
  Foundations of Software Engineering}, pp.\  1212--1216, 2026{\natexlab{g}}.

\bibitem[Li et~al.(2026{\natexlab{h}})Li, Li, and Li]{li2026defensible}
Zongwei Li, Wenkai Li, and Xiaoqi Li.
\newblock Defensible design for openclaw: Securing autonomous tool-invoking
  agents.
\newblock \emph{arXiv preprint arXiv:2603.13151}, 2026{\natexlab{h}}.

\bibitem[Liu et~al.(2004)Liu, Moore, Yang, and Gray]{liu2004investigation}
Ting Liu, Andrew Moore, Ke~Yang, and Alexander Gray.
\newblock An investigation of practical approximate nearest neighbor
  algorithms.
\newblock \emph{Advances in neural information processing systems}, 17, 2004.

\bibitem[Liu \& Song(2024)Liu and Song]{liu2024funredisp}
Yunqi Liu and Wei Song.
\newblock Funredisp: Reordering function dispatch in smart contract to reduce
  invocation gas fees.
\newblock In \emph{Proceedings of the 33rd ACM SIGSOFT International Symposium
  on Software Testing and Analysis}, pp.\  516--527, 2024.

\bibitem[Long et~al.(2025)Long, Wang, and Li]{Long2025From}
Xu~Long, Yishun Wang, and Xiaoqi Li.
\newblock From fomo3d to lottery dapp: Analysis of ethereum-based gambling
  applications.
\newblock \emph{arXiv}, 2025.

\bibitem[Luo et~al.(2026)Luo, Wang, Zhang, Zhang, Zhang, Zhao, Lin, Zhang, Liu,
  Tang, et~al.]{luo2026graph}
Haitong Luo, Fali Wang, Weiyao Zhang, Xianren Zhang, Zhiwei Zhang, Tianxiang
  Zhao, Minhua Lin, Jiahao Zhang, Hui Liu, Xianfeng Tang, et~al.
\newblock Graph-assisted large language models: A perspective on mitigating
  intrinsic limitations.
\newblock In \emph{Findings of the Association for Computational Linguistics:
  ACL 2026}, pp.\  18936--18955, 2026.

\bibitem[Ma et~al.(2024)Ma, Feng, Zeng, Lu, and Chen]{ma2024smart}
Jiarun Ma, Shiling Feng, Jiahao Zeng, Jia Lu, and Jie Chen.
\newblock Smart contract vulnerability detection based on prompt-guided
  chatgpt.
\newblock In \emph{2024 International Conference on Networking and Network
  Applications (NaNA)}, pp.\  321--326. IEEE, 2024.

\bibitem[Nemati et~al.(2025)Nemati, Abosata, and Butt]{nemati2025enhancing}
Nasrin Nemati, Nasr Abosata, and Usman~Javed Butt.
\newblock Enhancing ethereum smart contract security: A novel slither-based
  static analysis framework.
\newblock In \emph{2025 IEEE International Conference on Distributed Ledger
  Technologies (ICDLT)}, pp.\  1--7, 2025.

\bibitem[Pang et~al.(2024)Pang, Nol, and Heng]{pang2024chatgpt}
Samarnh Pang, Engheang Nol, and Kimkong Heng.
\newblock Chatgpt-4o for english language teaching and learning: Features,
  applications, and future prospects.
\newblock \emph{Available at SSRN 4837988}, 2024.

\bibitem[Pearlson et~al.(2024)Pearlson, Liu, Huang, George, Song, and
  Wang]{pearlson2024evaluating}
Joshua~Carter Pearlson, Xiaoyuan Liu, Chengsong Huang, Kripa~Ann George, Dawn
  Song, and Chenguang Wang.
\newblock Evaluating large language models in an emerging domain: a pilot study
  in decentralized finance.
\newblock In \emph{ICLR 2024 Workshop on Navigating and Addressing Data
  Problems for Foundation Models}, 2024.

\bibitem[Rao et~al.(2026)Rao, Rahman, and Sarker]{rao2026evaluating}
Ruicheng Rao, Mostafizur Rahman, and Md~Showaib Sarker.
\newblock Evaluating ethereum gas fee dynamics.
\newblock In \emph{SoutheastCon 2026}, pp.\  1--6. IEEE, 2026.

\bibitem[Ren \& Wei(2024)Ren and Wei]{ren2024sligpt}
Xiaolei Ren and Qiping Wei.
\newblock Sligpt: A large language model-based approach for data dependency
  analysis on solidity smart contracts.
\newblock \emph{Software}, 3\penalty0 (3):\penalty0 345--367, 2024.

\bibitem[Sahoo et~al.(2024)Sahoo, Meharia, Ghosh, Saha, Jain, and
  Chadha]{sahoo2024comprehensive}
Pranab Sahoo, Prabhash Meharia, Akash Ghosh, Sriparna Saha, Vinija Jain, and
  Aman Chadha.
\newblock A comprehensive survey of hallucination in large language, image,
  video and audio foundation models.
\newblock \emph{Findings of the association for computational linguistics:
  EMNLP 2024}, pp.\  11709--11724, 2024.

\bibitem[Sipos \& Szénási(2025)Sipos and Szénási]{Sipos2025Optimal}
Miklós Sipos and Sándor Szénási.
\newblock Optimal gas consumption in ethereum smart contracts: A targeted
  review of empirical results, design patterns and formal methods.
\newblock In \emph{2025 IEEE 25th International Symposium on Computational
  Intelligence and Informatics (CINTI)}, pp.\  517--522, 2025.

\bibitem[Song et~al.(2024)Song, Wei, Qu, and Wang]{song2024unveiling}
Han Song, Yihao Wei, Zhongche Qu, and Weihan Wang.
\newblock Unveiling decentralization: A comprehensive review of technologies,
  comparison, challenges in bitcoin, ethereum, and solana blockchain.
\newblock In \emph{2024 IEEE 6th Advanced Information Management, Communicates,
  Electronic and Automation Control Conference (IMCEC)}, volume~6, pp.\
  1896--1901. IEEE, 2024.

\bibitem[Sui et~al.(2025)Sui, Li, Zhang, Song, and Li]{sui2025bridging}
Yi~Sui, Chaozhuo Li, Chen Zhang, Dawei Song, and Qiuchi Li.
\newblock Bridging external and parametric knowledge: Mitigating hallucination
  of llms with shared-private semantic synergy in dual-stream knowledge.
\newblock In \emph{Proceedings of the 2025 Conference on Empirical Methods in
  Natural Language Processing}, pp.\  10845--10869, 2025.

\bibitem[Sun et~al.(2024)Sun, Wu, Xue, Liu, Wang, Xu, Xie, and
  Liu]{sun2024gptscan}
Yuqiang Sun, Daoyuan Wu, Yue Xue, Han Liu, Haijun Wang, Zhengzi Xu, Xiaofei
  Xie, and Yang Liu.
\newblock Gptscan: Detecting logic vulnerabilities in smart contracts by
  combining gpt with program analysis.
\newblock In \emph{Proceedings of the IEEE/ACM 46th International Conference on
  Software Engineering}, pp.\  1--13, 2024.

\bibitem[Ta \& Do(2024)Ta and Do]{ta2024study}
Minh~Thanh Ta and Tien~Quyet Do.
\newblock A study on gas cost of ethereum smart contracts and performance of
  blockchain on simulation tool.
\newblock \emph{Peer-to-Peer Networking and Applications}, 17\penalty0
  (1):\penalty0 200--212, 2024.

\bibitem[Tang et~al.(2026)Tang, Wang, and Zhang]{tang2026knowledge}
Pingzhi Tang, Yiding Wang, and Muhan Zhang.
\newblock Knowledge is not enough: Injecting {RL} skills for continual
  adaptation.
\newblock In \emph{Proceedings of the 64th Annual Meeting of the Association
  for Computational Linguistics}, volume~1, pp.\  11969--11997. Association for
  Computational Linguistics, 2026.

\bibitem[Tao et~al.(2024)Tao, Shen, Gao, Zhang, Li, Hua, Hu, Tao, and
  Ma]{tao2024llms}
Chongyang Tao, Tao Shen, Shen Gao, Junshuo Zhang, Zhen Li, Kai Hua, Wenpeng Hu,
  Zhengwei Tao, and Shuai Ma.
\newblock Llms are also effective embedding models: An in-depth overview.
\newblock \emph{arXiv preprint arXiv:2412.12591}, 2024.

\bibitem[Tharammal \& Nitnaware(2024)Tharammal and
  Nitnaware]{tharammal2024maximizing}
Rahul~Raghavan Tharammal and Prashant Nitnaware.
\newblock Maximizing efficiency in smart contract execution costs: Techniques
  for cost optimization on blockchain networks.
\newblock In \emph{2024 International Conference on Integration of Emerging
  Technologies for the Digital World (ICIETDW)}, pp.\  1--6, 2024.

\bibitem[Vaswani et~al.(2017)Vaswani, Shazeer, Parmar, Uszkoreit, Jones, Gomez,
  Kaiser, and Polosukhin]{vaswani2017attention}
Ashish Vaswani, Noam Shazeer, Niki Parmar, Jakob Uszkoreit, Llion Jones,
  Aidan~N Gomez, {\L}ukasz Kaiser, and Illia Polosukhin.
\newblock Attention is all you need.
\newblock \emph{Advances in neural information processing systems}, 30, 2017.

\bibitem[Wahab et~al.(2026)Wahab, Fadila, Razali, and Wong]{wahab2026gas}
Nur Haliza~Abdul Wahab, Juniardi~Nur Fadila, Nur~Faszha Razali, and Keng~Yinn
  Wong.
\newblock Gas-efficient smart contract design: Quantifying refactoring impact
  on {EVM} execution costs.
\newblock \emph{International Journal of Advanced Computer Science and
  Applications (IJACSA)}, 17\penalty0 (5), 2026.

\bibitem[Wang et~al.(2026)Wang, Li, Li, Li, Xie, and Zhang]{Wang2026LibScan}
Yishun Wang, Wenkai Li, Xiaoqi Li, Zongwei Li, Lei Xie, and Yuqing Zhang.
\newblock Libscan: Smart contract library misuse detection with iterative
  feedback and static verification.
\newblock \emph{Blockchain: Research and Applications}, pp.\  100521, 2026.

\bibitem[Wu et~al.(2025)Wu, Xing, and Li]{Wu2025Exploring}
Xiangfan Wu, Ju~Xing, and Xiaoqi Li.
\newblock Exploring vulnerabilities and concerns in solana smart contracts.
\newblock \emph{arXiv}, 2025.

\bibitem[Wu et~al.(2022)Wu, Li, Wang, Liu, Zhu, Zhu, and
  Hu]{Wu2022International}
Zhendong Wu, Shan Li, Bin Wang, Tianjian Liu, Yongsheng Zhu, Chenming Zhu, and
  Mingqing Hu.
\newblock Detecting vulnerabilities in ethereum smart contracts with deep
  learning.
\newblock In \emph{2022 4th International Conference on Data Intelligence and
  Security (ICDIS)}, pp.\  55--60, 2022.

\bibitem[Yaish et~al.(2024)Yaish, Qin, Zhou, Zohar, and
  Gervais]{yaish2024speculative}
Aviv Yaish, Kaihua Qin, Liyi Zhou, Aviv Zohar, and Arthur Gervais.
\newblock Speculative $\{$Denial-of-Service$\}$ attacks in ethereum.
\newblock In \emph{33rd USENIX security symposium (USENIX Security 24)}, pp.\
  3531--3548, 2024.

\bibitem[Zhang et~al.(2025{\natexlab{a}})Zhang, Dou, and Li]{Zhang2025DoS}
Chunyi Zhang, Fengjiao Dou, and Xiaoqi Li.
\newblock Dos attacks and defense technologies in blockchain systems: A
  hierarchical analysis.
\newblock \emph{arXiv}, 2025{\natexlab{a}}.

\bibitem[Zhang \& Zhang(2025)Zhang and Zhang]{zhang2025hallucination}
Wan Zhang and Jing Zhang.
\newblock Hallucination mitigation for retrieval-augmented large language
  models: A review.
\newblock \emph{Mathematics (2227-7390)}, 13\penalty0 (5):\penalty0 856, 2025.

\bibitem[Zhang et~al.(2025{\natexlab{b}})Zhang, Li, Cui, Cai, Liu, Fu, Huang,
  Zhao, Zhang, Chen, et~al.]{zhang2025siren}
Yue Zhang, Yafu Li, Leyang Cui, Deng Cai, Lemao Liu, Tingchen Fu, Xinting
  Huang, Enbo Zhao, Yu~Zhang, Yulong Chen, et~al.
\newblock Siren’s song in the ai ocean: A survey on hallucination in large
  language models.
\newblock \emph{Computational Linguistics}, 51\penalty0 (4):\penalty0
  1373--1418, 2025{\natexlab{b}}.

\bibitem[Zhang et~al.(2022)Zhang, Zhang, Li, and Smola]{zhang2022automatic}
Zhuosheng Zhang, Aston Zhang, Mu~Li, and Alex Smola.
\newblock Automatic chain of thought prompting in large language models.
\newblock \emph{arXiv preprint arXiv:2210.03493}, 2022.

\bibitem[Zhou et~al.(2024)Zhou, Hu, Yuan, Cui, Jin, Chen, Wu, Yuan, Jiang, Wu,
  et~al.]{zhou2024large}
Hao Zhou, Chengming Hu, Ye~Yuan, Yufei Cui, Yili Jin, Can Chen, Haolun Wu, Dun
  Yuan, Li~Jiang, Di~Wu, et~al.
\newblock Large language model (llm) for telecommunications: A comprehensive
  survey on principles, key techniques, and opportunities.
\newblock \emph{IEEE Communications Surveys \& Tutorials}, 2024.

\bibitem[Zhou et~al.(2025)Zhou, Lyu, and Li]{zhou2025blockchain}
Wenwen Zhou, Dongyang Lyu, and Xiaoqi Li.
\newblock Blockchain security based on cryptography: a review.
\newblock \emph{arXiv preprint arXiv:2508.01280}, 2025.

\end{thebibliography}
\bibliographystyle{iclr2026_conference}

\appendix

\end{document}